\pdfoutput=1
\documentclass[11pt]{article}
\usepackage{amssymb}
\usepackage{amsmath}
\usepackage[final]{acl}

\usepackage{times}
\usepackage{latexsym}

\usepackage[T1]{fontenc}
\usepackage[utf8]{inputenc}

\usepackage{microtype}

\usepackage{inconsolata}

\usepackage{graphicx}

\usepackage{booktabs}
\usepackage{multirow}
\usepackage{makecell}

\renewcommand{\paragraph}[1]{\noindent\textbf{#1}}

\title{Mathador-LM: A Dynamic Benchmark for Mathematical Reasoning \\ on Large Language Models}

\author{Eldar Kurtic\textsuperscript{*}\\
  ISTA \& Neural Magic, Inc.\\
  \texttt{eldar.kurtic@ist.ac.at} \\\And
  Amir Moeini\textsuperscript{*}\\
  ISTA\\
  \texttt{amir.moeini@ist.ac.at} \\\And
  Dan Alistarh \\
  ISTA \& Neural Magic, Inc. \\
  \texttt{dan.alistarh@ist.ac.at} \\
}

\begin{document}
\maketitle
\def\thefootnote{*}\footnotetext{Equal contribution}\def\thefootnote{\arabic{footnote}}
\begin{abstract}
We introduce Mathador-LM, a new benchmark for evaluating the mathematical reasoning on large language models (LLMs), combining ruleset interpretation, planning, and problem-solving. This benchmark is inspired by the Mathador game, where the objective is to reach a target number using basic arithmetic operations on a given set of base numbers, following a simple set of rules.
We show that, across leading LLMs, we obtain stable average performance while generating  benchmark instances \emph{dynamically}, following a target difficulty level.
Thus, our benchmark alleviates concerns about test-set leakage into training data, an issue that often undermines popular benchmarks.
Additionally, we conduct a comprehensive evaluation of both open and closed-source state-of-the-art LLMs on Mathador-LM. Our findings reveal that contemporary models struggle with Mathador-LM, scoring significantly lower than average 3rd graders. This stands in stark contrast to their strong performance on popular mathematical reasoning benchmarks. The implementation of Mathador-LM benchmark is available at \href{https://github.com/IST-DASLab/Mathador-LM}{\texttt{github.com/IST-DASLab/Mathador-LM}}.
\end{abstract}

\section{Introduction}

\begin{figure*}[t]
  \centering
  \includegraphics[width=\textwidth]{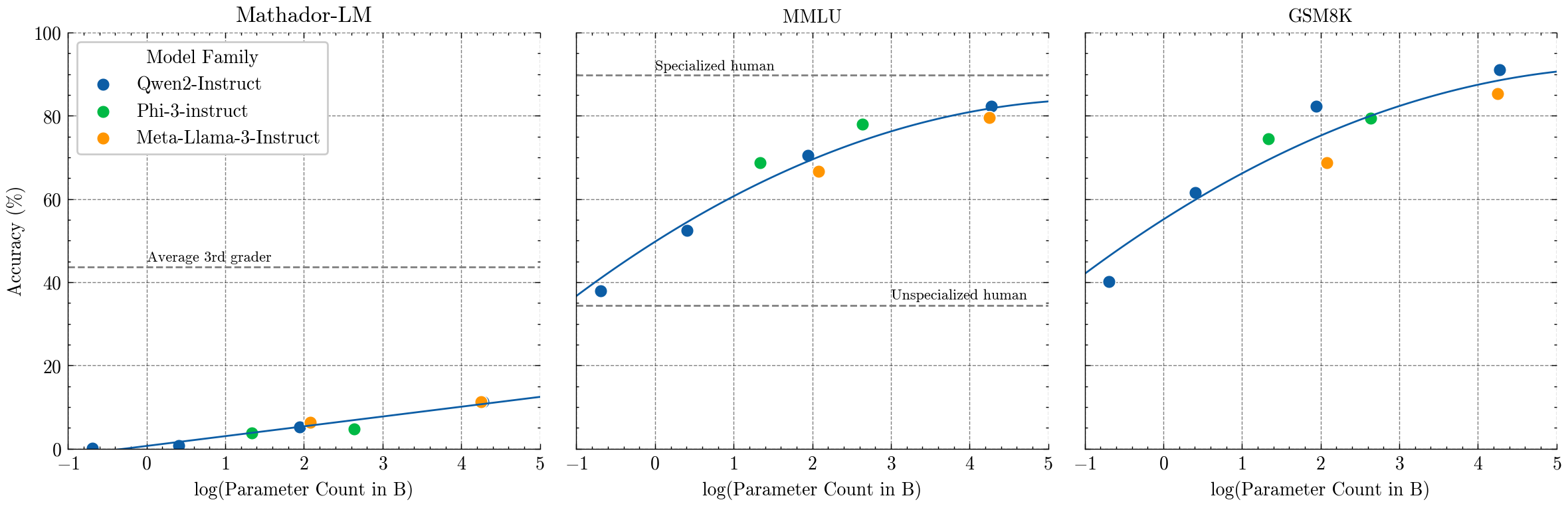}
  \caption{Comparative results on Mathador-LM, MMLU, and GSM8k, across the Llama-3-Instruct (8B and 70B), Phi-3-Instruct (small and medium), and Qwen2-Instruct model families. Interpolation lines show very high scores and clear saturation on MMLU and GSM8k at or beyond the level of specialized humans, whereas on Mathador-LM contemporary models are significantly below the average 3rd grader. MMLU and GSM8K results are obtained from \citet{open-llm-leaderboard}, \citet{mmlu}, and \citet{qwen}.
  }
  \label{fig:param_corr}
\end{figure*}

The ability of large language models (LLMs) to approach non-trivial tasks involving both information retrieval and mathematical reasoning has led to significant research interest in evaluating these properties. 
Yet, the popularity of reasoning benchmarks, such as the often-used Grade-School Math (GSM)~\cite{cobbe2021training} or MATH~\cite{hendrycks2021measuring} datasets, is leading to performance saturation (see Figure~\ref{fig:param_corr}), and can potentially lead to training set contamination. 
Thus, there is a stringent need to develop new strong benchmarks to evaluate LLM reasoning. 

We address this by proposing \emph{Mathador-LM}, a new benchmark for examining the mathematical reasoning properties of LLMs. At a high level, Mathador-LM follows the popular Mathador mathematical game~\cite{mathador}, in which a human player is given five base numbers together with a target number, and has to provide a series of calculations, each using one of the four basic arithmetic operations, which result in the target number.\footnote{Our game formulation follows the mathematical game organized in France for students between the 3rd and 8th grades, to which more than 10'000 pupils participated in 2023.} Each base number can only be used once, and solutions are scored on the number of  operations used---a ``perfect'' solution uses each basic operation and each base number exactly once. 

We define and implement Mathador-LM following the framework for few-shot evaluation of language models~\cite{gao2021framework}, and evaluate leading open and closed LLMs such as Llama~\cite{llama3}, and Qwen2~\cite{qwen}, as well as Claude~\cite{anthropic2023claude} and GPT3.5/4~\cite{achiam2023gpt}. 
Our key observations are: 

\begin{itemize}
    \item \emph{Mathador is a hard benchmark for LLMs}: state-of-the-art open and closed models score below 15\% on average, which is  significantly below the mean of 43.7\% across 3rd-grade students in 2023~\cite{ResultatsConcoursMathador}. 
    \item We observe clear correlations between model size and game performance, where models below 3B parameters obtain negligible accuracy, state-of-the-art models in the 7-8B range obtain scores of 5-7\%, and 70-72B models reach the top scores of 10-15\%, together with Claude-Opus. Remarkably, GPT4 and Claude-Haiku models both obtain scores below 7\%. 
    \item We introduce a notion of difficulty to the classic Mathador game~\cite{mathador} to enhance the benchmark's robustness and future-proof it. We then perform a detailed analysis of the performance breakdown and failure modes at the base difficulty level, which aligns with the one used for human evaluation.
    \item Importantly, Mathador-LM has the property that model performance is \emph{stable across randomly-generated problem instances of the same difficulty}. Thus, we can generate one-time \emph{dynamic} instances of similar difficulty, preventing ``over-fitting.'' 
\end{itemize}

Our results are especially relevant in the context of recent work by~\citet{yang2023rethinking} and ~\citet{gunasekar2023textbooks} raising concerns about contamination across popular benchmarks used to evaluate the performance of LLMs. Their findings span three different axes: 1) existing decontamination techniques often fail to identify problematic samples, 2) synthetic data generated by closed-source models (e.g., GPT-3.5/4~\cite{achiam2023gpt}) exhibits subtle test-set contamination, and 3) popular open-source datasets (e.g., RedPajama~\cite{together2023redpajama}, StarCoder~\cite{li2023starcoder}, The Stack~\cite{kocetkov2022stack}, FLAN CoT~\cite{longpre2023flan}) are also contaminated to varying degrees, ranging from 0.5\% to 19\%~\cite{yang2023rethinking}. 
This evidence, together with the fact that performance on the few standard benchmarks~\cite{cobbe2021training, hendrycks2021measuring}  for mathematical reasoning is rapidly saturating\footnote{For instance, the best achieved accuracy on GSM at the time of writing is already of 97.1\%~\cite{zhong2024achieving}.}, as described in Figure~\ref{fig:param_corr},    necessitates enhancing our existing evaluation protocols and significantly improving the decontamination of existing datasets with static benchmarks. 

We propose an alternative pathway towards reliable examination of LLM performance via \emph{dynamic, one-time benchmarks} that mitigate contamination by being created \emph{on-the-fly, independently} for each evaluation run. 
 Mathador-LM satisfies these properties: given its nature, the benchmark can be programmatically generated and verified, making it ideally suited for fresh, one-time evaluations of LLMs. This approach mitigates issues such as test-set leakage into training data and provides a reliable method to evaluate closed-source models, even in the absence of detailed information about their training data. Moreover, results reveal interesting trends across different model families and sizes, and allowing to isolate model proficiency across instruction-following, mathematical reasoning, planning, and combinatorial search. 

\section{The Mathador-LM Benchmark}
\label{sec:definition}

\begin{figure}[h!]
  \centering
  \includegraphics[width=0.4\textwidth]{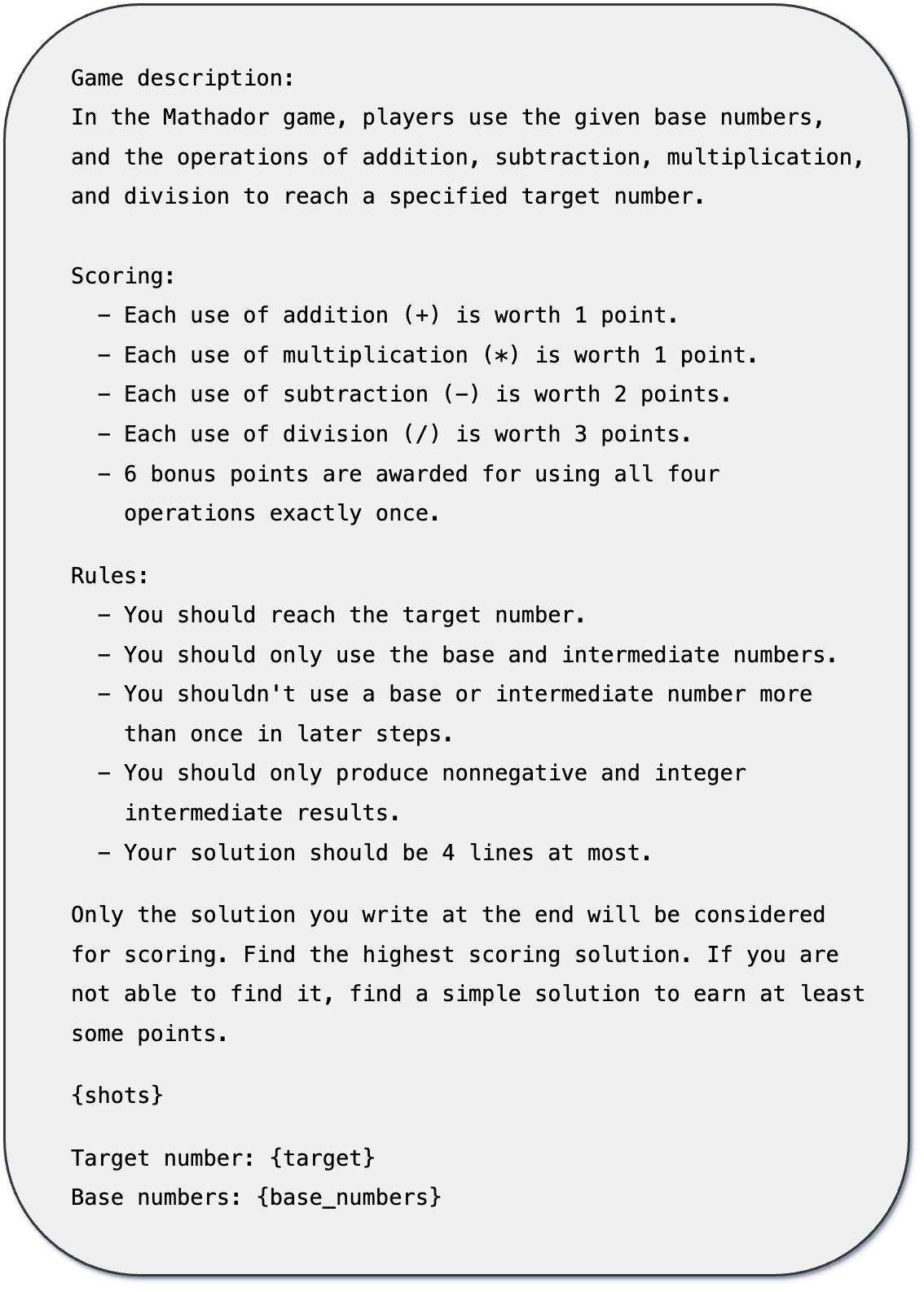}
  \caption{The prompt for Mathador-LM benchmark.}
  \label{fig:prompt}
\end{figure}

\begin{figure}[h!]
  \centering
  \includegraphics[width=0.35\textwidth]{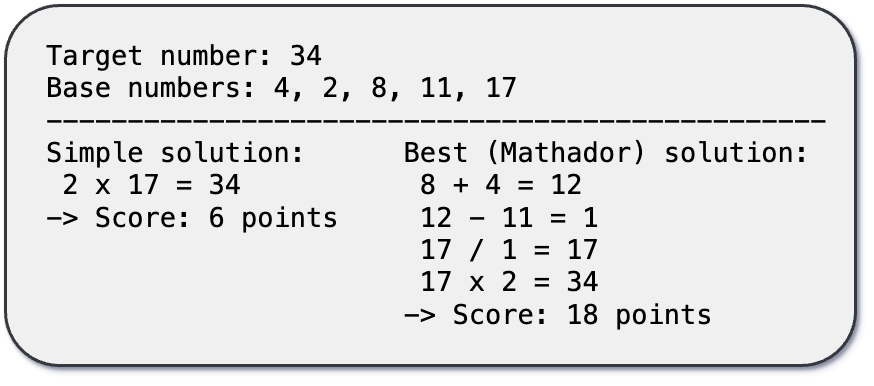}
  \caption{An example problem demonstrating both simple and best (Mathador) solutions.}
  \label{fig:example}
\end{figure}

The informal definition of the Mathador-LM game we use is provided in Figure~\ref{fig:prompt}, which coincides with the prompt we provide to the LLM in the default version of the game. In Table~\ref{tab:mathador_scoring} we present the scoring system for the benchmark. An example instance of the benchmark is provided in Figure~\ref{fig:example}, together with basic and ``optimal''  solutions. 

\begin{figure*}[t]
  \centering
  \includegraphics[width=0.85\textwidth]{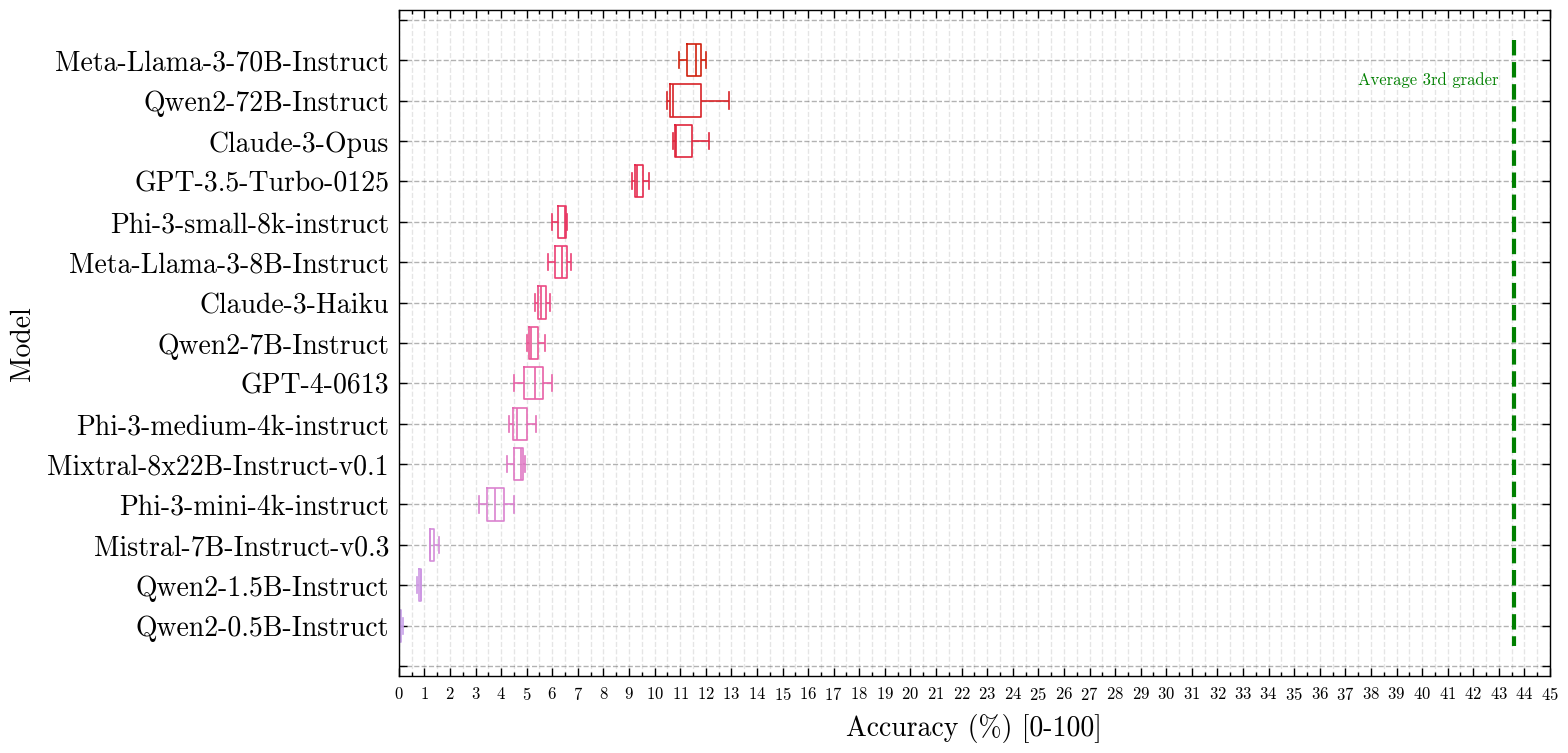}
  \caption{Detailed results on Mathador-LM across open and closed models, including confidence intervals. Experiments performed in June 2024.}
  \label{fig:gantt}
\end{figure*}

\paragraph{Formal Definition.}
Given a set of operands \mbox{$A = \{a_i \in \mathbb{N}| 1 \leq i \leq 5\}$} and target value $t \in \mathbb{N}$,
let $P \in \{S! | S\in \mathcal{P}(A)\}$ be a permutation of a subset of operands and 
define the set of expressions
\small
\begin{equation*}
    \mathcal{E}_P=\Big\{(P^c, O)| P^c \in C(P), O \in \{+, \times, -, \div\}^{|P|}\Big\}
\end{equation*}
\normalsize
where $C(P)$ is the set of all legal \textit{parenthesization} of $P$. Consequently the set of all expressions $\mathcal{E} = \bigcup_{P}\mathcal{E}_P$.
Each expression $E \in \mathcal{E}$ has the value $\operatorname*{val}(E)$ which is derived by associating the $i$th opening parenthesis in $P^c$ with the operator $O_i$.
 Given the score function $s: \mathcal{E} \to \mathbb{N}$ we are looking for
 $E^* = \operatorname*{argmax}_{E \in \mathcal{E}}s(E) \; \text{s.t.} \; \operatorname*{val}(E)=t.$

Each expression $E$ can be represented in an expanded form $\operatorname*{repr}(E)$ by writing the evaluation of each parenthesis when both of its nested values have been evaluated. For instance, $\operatorname*{repr}(E)$ of 
$E = \Big( ((17,((8,4),11)),2), (\times, \div, -, +) \Big)$ is the Mathador solution illustrated in Figure~\ref{fig:example}. 
In Mathador-LM we use $\operatorname*{repr}(E)$ as the representation since it is more human-readable and Table~\ref{tab:mathador_scoring} for scoring. The \textit{accuracy} of expression $E$ is defined as $s(E)/s(E^*)$.

\paragraph{Difficulty Measure.}
For a specific set of operands,
$
E_t = \{ E \in \mathcal{E} | \operatorname*{val}(E) = t, s(E) > 0\}
$
is the set of all \textit{solutions} for target $t$. We define the difficulty measure of target $t$ as 
${\sum_{E \in E_t} s(E)} / 
    {|E_t|^2},$ following the intuition that instances with few but higher-scoring solutions are harder.

\begin{table}[h]
\centering
\caption{Scoring system for Mathador-LM benchmark. The Mathador Bonus refers to the optimal solution, achieved by using all five base numbers and each of the four operators exactly once.}
\label{tab:mathador_scoring}
\small{
    \begin{tabular}{ll}
    \toprule
    \textbf{Category} & \textbf{Points} \\
    \midrule
    Target number reached & 5 points \\
    \textbf{Operators} & \\
    \quad Addition & 1 point \\
    \quad Multiplication & 1 point \\
    \quad Subtraction & 2 points \\
    \quad Division & 3 points \\
    Mathador Bonus & 6 points \\
    \textbf{Invalid Solutions} & \\
    \quad Target number not reached & 0 points \\
    \quad Reuse of numbers & 0 points \\
    \quad Negative numbers & 0 points \\
    \quad Non-integer numbers & 0 points \\
    \bottomrule
    \end{tabular}
}
\end{table}

\section{Model Evaluations}

\paragraph{Evaluation Setup.} A dataset of Mathador-LM problems is generated for each model evaluation by sampling the operand dataset $A$ based on the official rules~\cite{mathador} and then sampling from possible targets $\{t | \exists E \in \mathcal{E} \text{ s.t. } \operatorname*{val}(E) = t\}$ based on the desired difficulty distribution. The restrictions for base numbers are closely following the official rules of the human game, so that results we obtain with LLMs are directly comparable to human results. This means that base numbers are sampled as integers, uniformly at random from the following ranges: $n_1 \in [1,4]$, $n_2 \in [1,6]$, $n_3 \in [1,8]$, $n_4 \in [1,12]$, and $n_5 \in [1,20]$.
The prompt in Figure~\ref{fig:prompt} is populated based on a newly generated problem set to get the final prompt. The model's generated answer to the prompt is parsed to get the solution block which is then scored. In addition to the prompt shown in Figure~\ref{fig:prompt} we have experimented with everything from concise descriptions to extensive, detailed explanations. In terms of few-shot prompts, we have also tried quite a few approaches to present the model with step-by-step solutions. Some prompts were designed and tested following an error analysis that helped identify the most common mistakes made by the models. For instance, we developed specific prompts after observing that the majority of errors involved the use of illegal operands. These prompts explicitly specify the set of permissible operands at each step. Unfortunately, despite these additional instructions, we have not seen any noticeable accuracy gains. 

Figure~\ref{fig:gantt} presents evaluations on several popular open and closed models. 
We observe that small models ($\leq 3B$) and Mistral-7B tend to perform below $<2\%$ average accuracy (0.36 points per instance, on average), meaning that they reach a correct solution (worth $\geq 6$ points) less than 6\% of the time. Surprisingly, well-performing medium models such as Qwen2-7B, Llama-3-8B, and \mbox{Phi-3-medium} perform on par with GPT 3.5 and GPT4, as well as Claude-Haiku (5 to 7\%), at a level corresponding to reaching a correct solution less than 20\% of the time. 
Further, we observe a higher tier for 70B models and Claude-Opus, which reach similar $\sim 12\%$ performance. In Appendix~\ref{app:score_distribution} we expand our analysis, and detail the score distribution across models. In Appendix~\ref{app:multi_attempt} we investigate how allowing multiple attempts to solve each question affects the accuracy of LLMs. 

\paragraph{Stability.} 
A reliable benchmark must be reproducible,  which is why most benchmarks are \emph{static}. Table~\ref{tab:difficulty_stability} shows that we can obtain consistent scores on Mathador-LM even when we \emph{dynamically re-generate} the benchmark, by sampling instances with a similar difficulty mix. The \textit{easy}, \textit{medium}, and \textit{hard} datasets are taken from the beginning, middle, and end of the sorted list of targets, based on difficulty (see Section~\ref{sec:definition}). The \textit{mixed} dataset contains equal fractions from each type. 

\begin{table}[h!]
\centering
\caption{Stability across 5 evaluations of LLama-3-70B-Instruct on datasets of varying sizes and difficulties. Observe that the performance on the standard ``mixed'' benchmark is very stable across number of samples.}
\label{tab:difficulty_stability}
\small{
    \begin{tabular}{ccc}
    \toprule
    \textbf{\# Samples} & \textbf{Difficulty} & \textbf{Accuracy (\%)} \\ \hline
    100 & mixed & 12.3 ± 1.7 \\
    250 & mixed& 11.8 ± 1.1 \\ 
    500 & mixed& 11.5 ± 0.5 \\ \hline
    \multirow{4}{*}{1000} & easy & 15.1 ± 0.8 \\ 
    & medium& 12.1 ± 0.6 \\
    & hard& \phantom{0}4.3 ± 0.2 \\
    & mixed& 11.3 ± 0.5 \\ \hline
    1500 & mixed & 12.0 ± 0.5 \\ 
    \bottomrule
    \end{tabular}
}
\end{table}

\paragraph{Impact of Number of Shots.}
We investigate whether increasing the number of ``shots'' in the few-shot evaluation setup helps performance on Mathador-LM, as few-shot prompting~\cite{brown2020language} is known to enhance in-context learning abilities of LLMs~\cite{wei2022emergent}. 
We report results in Table~\ref{tab:shots_ablations}.
Surprisingly, for Mathador-LM, we found that two shots are sufficient to grasp the  formatting and evaluation flow. Further increasing of this number only marginally improves results. In Appendix~\ref{app:text_decoding} we further explore how the results are affected by different text-generation (decoding) strategies, such as greedy~\citep{radford2019language} and nucleus sampling~\citep{holtzman2019curious}.

\begin{table}[h]
\centering
\caption{Impact of the number of shots on the evaluation of Llama-3-70B-Instruct on Mathador-LM.} 
\label{tab:shots_ablations}
\setlength{\tabcolsep}{2pt}
\small{
    \begin{tabular}{ccccc}
    \toprule
    \textbf{\# shots} & 2 & 5 & 10 & 20 \\ \hline
    \textbf{\makecell{Accuracy\\(\%)}} & 13.1 ± 0.6 & 13.9 ± 0.7 &  14.25 ± 0.6 &  14.34 ± 0.9 \\ 
    \bottomrule
    \end{tabular}
}

\end{table}

\paragraph{Errors Analysis.}
In Table~\ref{tab:errors} we present a breakdown of the errors that LLMs make when evaluated on Mathador-LM benchmark, categorized into four types: Formatting, Calculation, Missed Target, and Illegal Operand. Formatting errors occur when the model fails to adhere to the expected format for the intermediate steps. Calculation errors happen when the model makes mistakes in basic arithmetic operations while generating a solution. Missed Target refers to cases where the model correctly follows the expected format and performs all calculations accurately but arrives at a different target number than the expected solution. Illegal Operand errors arise when the model uses a number not included in the set of allowed values to produce the final solution. This can happen if a number is either not part of the base numbers provided or has already been used in previous steps.

The results in Table~\ref{tab:errors} highlight that the most significant challenges faced by the model are related to the use of illegal operands, which collectively make up over 60\% of the errors. This indicates that existing models still struggle even with moderate reasoning abilities. (This complements the recent findings of~\citet{nezhurina2024alice}.)  
To address the most common error made by LLMs (Illegal Operand), we augmented our prompting strategy to explicitly show the model the set of allowed operands at each step of the calculation process. Surprisingly, this \emph{did not} improve results.

\begin{table}[h]
\centering
\caption{Error types of instruction-following models on Mathador-LM, in percentages.} 
\label{tab:errors}
\setlength{\tabcolsep}{2pt}
\small{
    \begin{tabular}{ccccc}
    \toprule
     & \makecell{Formatting\\Error} & \makecell{Calculation\\Error} & \makecell{Missed\\Target} & \makecell{Illegal\\Operand} \\ \hline
    Qwen2-7B & 5.5 & 20.9 & \phantom{0}6.8 & 66.8 \\ 
    Llama-3-8B & 0.3 & 17.3 & \phantom{0}7.1 & 75.3 \\ 
    Llama-3-70B & 0.9 & \phantom{0}3.1 & 32.5 & 63.5 \\ 
    \bottomrule
    \end{tabular}
}
\end{table}
\vspace{-.5em}
\section{Limitations}
\vspace{-.5em}

We introduced a new challenging LLM mathematical reasoning benchmark. Our benchmark is dynamic, as it can be generated on-the-fly, mitigating the risks of test-set leakage and overfitting. 
The current setup can be easily extended to vary difficulty levels by, for example, adjusting the ranges of base numbers, or the total number of operands.

By design, Mathador-LM  is limited to a search-based mathematical task, which has been linked to both conceptual and procedural skills~\cite{mathador}. 
Another limitation we plan to investigate in future work are more advanced prompting techniques, which might alleviate the relatively low LLM performance on this task. 
Additionally, we plan to explore supervised fine-tuning strategies.

\section*{Acknowledgments}

The authors would like to express their gratitude to TogetherAI for providing the computational resources that made this research possible.

\bibliography{custom}

\appendix

\section{Score Distribution}
\label{app:score_distribution}
\begin{figure*}[h]
  \includegraphics[width=\textwidth]{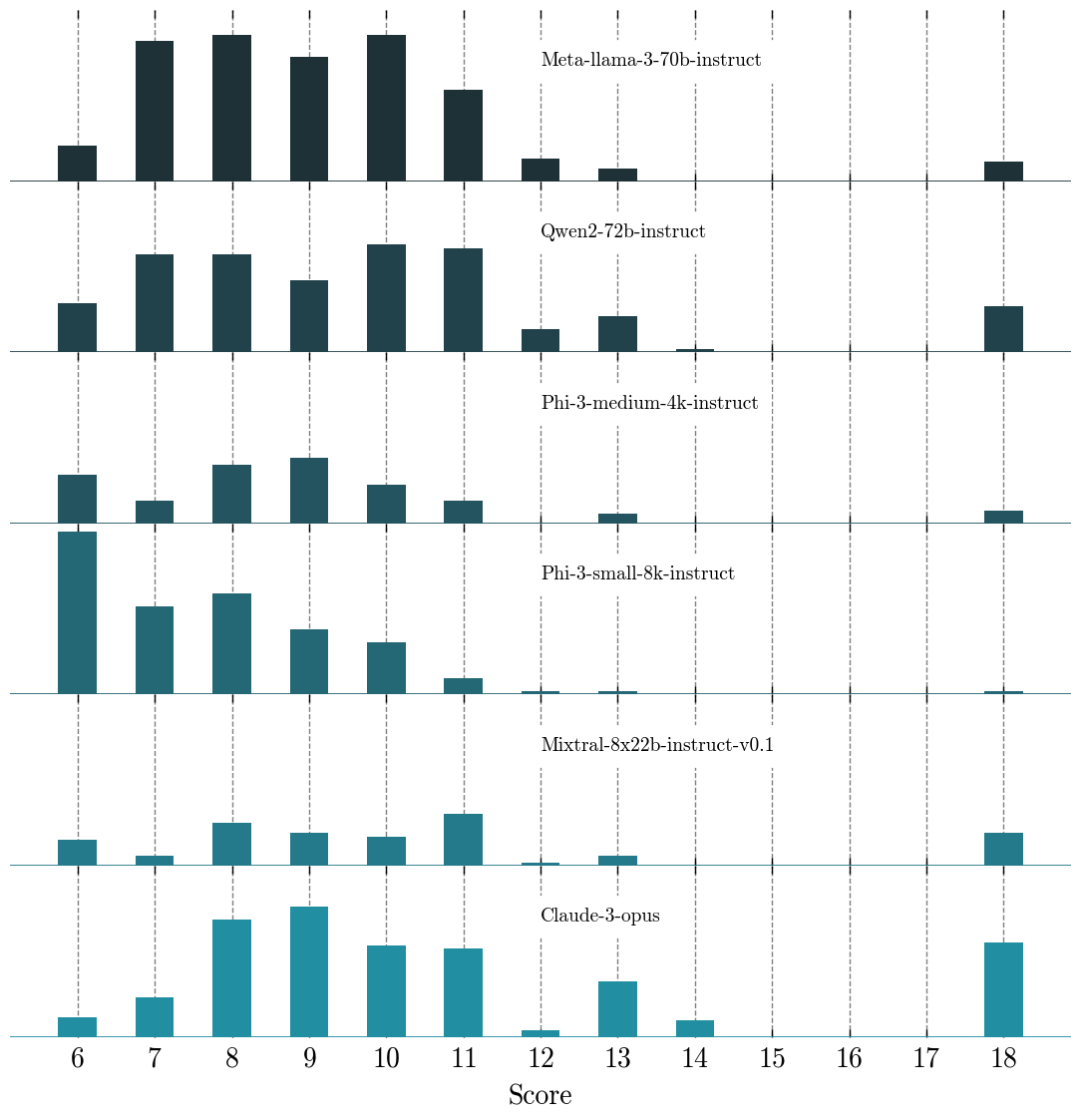}
  \caption{
  Distribution of scores for several models showing low correlation of higher overall performance with number of high scoring solutions.
  }
  \label{fig:scoredist}
\end{figure*}

Models are instructed that only their last answer will be scored, and there is no obvious strategy for reaching a more complicated and higher scoring answer from a lower scoring one, as this is part of the task. Consequently, it is natural that even similarly performing models may have quite different score distributions as they may aim to  obtain answers with different complexity levels (e.g., one may aim to obtain only highest-scoring answers, but may fail to obtain one more often than if simply aiming to reach the target). Figure~\ref{fig:scoredist} shows the score distribution for several low and high performing models. For instance, it is interesting to observe that Claude-3-opus outputs several times more max-scoring solutions than Llama-3-70b-instruct, while the models score about the same on average, based on Figure~\ref{fig:gantt}, or that Phi-3-small focuses on obtaining simple answers correct (just reaching the target, but not focusing on reaching high scores), which has resulted in a \emph{higher overall performance} relative to Phi-3-medium, which produces higher-scoring solutions.
x
\section{Evaluation with Multiple Attempts}
\label{app:multi_attempt}
In this section, we analyze the impact of allowing multiple attempts per question. During evaluation, we permit the model up to $K$ attempts for each question. For $K = 5$, we observe a noticeable improvement in accuracy; however, it still falls short of being competitive with human performance. The results are presented in Table~\ref{tab:multiple}.

\begin{table}[h]
\centering
\caption{Results with multiple attempts allowed to solve each question in Mathador benchmark.} 
\label{tab:multiple}
\small{
    \begin{tabular}{cccc}
    \toprule
     Model & 1 attempt & 5 attempts & Gain \\ \hline
    Llama-3-8B & 6.32 $\pm$ 0.47 & 8.15 $\pm$ 0.12 & 29\% \\
    Llama-3-70B & 11.52 $\pm$ 0.52 & 13.70 $\pm$ 0.58 & 19\% \\
    \bottomrule
    \end{tabular}
}
\end{table}

\section{Text Generation Strategies}
\label{app:text_decoding}
Given that the nature of Mathador-LM benchmark is based on generating text to arrive at a solution, we investigate whether different decoding methods for language generation have any effect on the results. Therefore we consider both, the simple greedy decoding~\citep{radford2019language} and the more advanced nucleus sampling~\citep{holtzman2019curious}. We conduct an extensive search, exploring all possible combinations of \textit{temperature} (0.0, 0.3, 0.5, 0.7, 0.9) and \textit{Top-p} (0.1, 0.3, 0.5, 0.7, 1.0) hyper-parameters. As can be seen from Table~\ref{tab:decodings}, the results are not affected by choices of different text-generation strategies.

\begin{table}[h]
\centering
\caption{Results with Llama-3-70B-Instruct on Mathador-LM benchmark under different text decoding techniques, evaluated across three few-shot configurations.} 
\label{tab:decodings}
\small{
    \begin{tabular}{cccc}
    \toprule
     & 2-shots & 5-shots & 20-shots \\ \hline
    Greedy & 12.8 ± 0.5	& 13.9 ± 0.1 & 14.2 ± 1.1 \\ 
    Nucleus & 13.1 ± 0.6 & 13.8 ± 0.7 & 14.2 ± 0.9 \\
    \bottomrule
    \end{tabular}
}
\end{table}

\end{document}